# Analytical Solution for Inverse Kinematics

Serdar Kalaycioglu[1,2], Anton de Ruiter[1], Ethan Fung[2], Harrison Zhang[2], Haipeng Xie[2]

[1]Department of Aerospace Engineering, Toronto Metropolitan University, Toronto, ON, Canada-M5B 2K3
[2]Dr Robot Inc., 44 East Beaver Creek Rd, Suite 8, Richmond Hill, ON, Canada-L4B 1G8

*Abstract*— *This paper introduces a closed-form analytical solution for the inverse kinematics (IK) of a 6 Degrees of Freedom (DOF) serial robotic manipulator arm, configured with six revolute joints and utilized within the Lunar Exploration Rover System (LERS). As a critical asset for conducting precise operations in the demanding lunar environment, this robotic arm relies on the IK solution to determine joint parameters required for precise end-effector positioning, essential for tasks such as sample collection, infrastructure assembly, and equipment deployment. By applying geometric principles, the proposed method offers a highly efficient and accurate approach to solving the IK problem, significantly reducing computational demands compared to traditional numerical methods. This advancement not only enhances real-time operational capabilities but is also optimized for space robotics, where precision and speed are critical. Additionally, the paper explores the integration of the LERS robotic system, underscoring the importance of this work in supporting autonomous lunar exploration within the ARTEMIS program and future missions.*

## I. INTRODUCTION

Space exploration has evolved from a speculative endeavor to a vital driver of scientific advancement, with the Moon now a focal point for pushing the boundaries of space technology. At the heart of this mission lies the development of Lunar Exploration Rover Systems (LERS), engineered to propel both lunar scientific discovery and the technological infrastructure crucial for future space missions. These rovers not only serve as platforms for advanced scientific instruments but also function as testing grounds for innovations that will define the future of lunar and planetary exploration. The capability of LERS to navigate and interact with the Moon's rugged terrain is pivotal to mission success, particularly in environments demanding energy efficiency, adaptability, and autonomous operation. [1-8].

The **ARTEMIS program**, an international collaboration led by NASA, with contributions from the European Space Agency (ESA), the Canadian Space Agency (CSA), and the Japan Aerospace Exploration Agency (JAXA), is a pivotal initiative aimed at returning humans to the Moon and establishing a sustainable presence. The ARTEMIS program seeks to conduct detailed scientific exploration of the lunar surface and test the technologies needed for future missions to Mars. LERS and similar robotic systems will be crucial for supporting both crewed and uncrewed missions, enabling precise manipulation tasks such as deploying infrastructure, gathering scientific data, and managing lunar resources [9].

As lunar exploration technologies continue to advance, the focus has shifted beyond just mobility to include precise robotic manipulation. This capability is essential for tasks such as assembling lunar habitats, handling materials, and conducting in-depth scientific analyses. The robotic manipulator arms mounted on systems like LERS require precise control to perform these operations effectively, and this is where **inverse kinematics (IK)** becomes vital. Accurate IK solutions allow the robotic arm to position its end-effector correctly in relation to its base, ensuring that it can perform delicate tasks such as assembling infrastructure, managing space debris, or manipulating lunar samples. The implementation of IK in space-based robotics, particularly in the context of LERS and the ARTEMIS program, represents a critical component in ensuring that these systems can operate with the precision needed for future lunar missions.

In robotics, IK is essential for determining the required joint movements to place a manipulator's end-effector in a desired position and orientation, which is fundamental for tasks like pick-and-place operations, assembly, and welding [10]. IK plays a vital role in motion planning, where robots need to navigate complex environments while adhering to constraints such as obstacle avoidance, minimizing energy consumption, or achieving specific postures [11]. For example, industrial robots often rely on IK to enable precise and efficient movements during manufacturing processes, allowing automation in tasks requiring high accuracy (Schilling, 1990) [12]. Additionally, in medical robotics, IK is indispensable for controlling surgical robots that perform delicate procedures in confined spaces, where precision and dexterity are paramount [13].

IK is a critical concept not only in robotics but also in numerous other fields, such as computer graphics, animation, biomechanics, and even virtual reality. In 3D animation and video games, IK allows for more natural character movements by automating the computation of joint angles required for a character's limbs to reach a specific target [14]. This technique is particularly important for simulating lifelike human and animal motion, ensuring that characters can move and interact with their environment in a realistic manner [15]. In biomechanics, IK is used to study human movement by calculating joint angles based on external markers, providing valuable insights into the mechanics of walking, running, and performing specialized tasks. This information is crucial for designing prosthetics, improving athletic performance, and developing ergonomic tools [16]. Moreover, in virtual and augmented reality, IK algorithms are employed to accurately track users' body movements and render them in virtual environments, thereby enhancing immersion and interaction [17,18].

IK has been extensively studied, with methods broadly categorized into **analytical**, **numerical**, **hybrid**, and **machine learning-based** approaches.

**Analytical methods** are widely favored in simpler robotic systems because they provide closed-form solutions, which are computationally efficient and exact. Early work by Pieper (1968) [19] solved the IK problem for 6-DOF robots with spherical wrists, leading to significant advances in industrial robots like the PUMA and Intelledex 660T [20,12]. Kang (1993) and Zaplana and Basanez (2018) further explored closed-form solutions for robots with specific kinematic structures, emphasizing their computational advantages [21,22]. These methods rely heavily on geometric and algebraic properties of the manipulator, making them ideal for systems with simple kinematics [23]. However, as noted by Paul (1979), analytical solutions become challenging to derive for manipulators with more complex kinematic chains, limiting their practical applications.

To address the limitations of analytical solutions, **numerical methods** have been developed. These methods iteratively converge to a solution by minimizing the error between the desired and current end-effector positions. Jacobian-based techniques, such as the **Jacobian transpose** and **pseudoinverse** methods, are foundational in this area [24,25]). The **Damped Least Squares (DLS)** method is a widely used extension that improves stability near singularities [24,26]). Numerical methods, such as **Cyclic Coordinate Descent (CCD)**, offer flexibility but can suffer from issues such as high computational cost and slow convergence [18,27]. Additionally, Hollerbach (1985) and Kreutz-Delgado and Seraji (1992) explored torque optimization in numerical IK, particularly for redundant manipulators [28, 29]. Hybrid methods that blend analytical and numerical approaches have also become popular due to their ability to exploit the strengths of both [30].

**Hybrid methods** are particularly effective in complex environments, such as space robotics, where real-time control and robustness against singularities are critical. For example, the Space Station Remote Manipulator System (SSRMS) and the European Robotic Arm (ERA) use hybrid IK methods for high adaptability [31-33]. In space environments, hybrid solutions such as joint angle parameterization, as discussed by Xu et al. (2014) and Qian et al. (2020), are applied to redundant manipulators to achieve better control [34,35]. These methods integrate the computational efficiency of analytical approaches with the flexibility of numerical methods, as demonstrated by Han et al. (2022) and Fu et al. (2022a, 2022b) [36-38].

One major challenge in IK is **handling redundancy**, which arises when a manipulator has more than six DOF, allowing multiple joint configurations for a given end-effector pose. Kreutz-Delgado et al. (1993) and Hollerbach (1985) discussed the optimization of joint torques in redundant systems using null-space optimization techniques [39,28]. Such redundancy is crucial in systems like the SSRMS, where secondary objectives such as obstacle avoidance or minimizing joint torques can be achieved simultaneously [26,40]. Methods such as joint angle parameterization (Xu et al., 2014 [33]) and optimization-based solutions (Sinha and Chakraborty, 2019) have been developed to resolve redundancy efficiently [16,41]].

More recently, **machine learning** approaches have gained traction in solving the IK problem. Neural networks can be trained to approximate the inverse mapping from Cartesian space to joint space, offering potential solutions that handle redundancy and singularities intuitively [13,14]). However, as noted by Flores-Abad and Fu (2015,2020), machine learning approaches often require extensive training data and can suffer from high computational costs, limiting their applicability in real-time control [42,43]. Jin et al. (2020) and Tian et al. (2021) explored the use of optimization techniques such as **particle swarm optimization** to improve learning efficiency, but challenges remain in achieving both computational speed and accuracy [17,44]. Hybrid methods combining machine learning with traditional numerical techniques, such as those proposed by Broyden (1965, 1970), aim to reduce the computational load while maintaining flexibility and adaptability [45,46].

As robotic applications expand, from industrial automation to space exploration, the demand for more precise and efficient inverse kinematics (IK) solutions continues to grow. **Closed-form analytical methods** remain highly important due to their exactness and superior computational efficiency, particularly in structured systems. These methods provide precise solutions without iterative computations, making them ideal for real-time applications where speed and accuracy are critical. While numerical and hybrid methods are often employed for handling complex and redundant manipulators, they are prone to computational delays and approximation errors, which can limit their effectiveness in high-speed, precision-dependent tasks. Although machine learning approaches show promise for complex systems, they still require significant development to meet the stringent real-time and precision demands of advanced robotic applications. **Analytical methods**, by contrast, offer a robust and reliable approach that can be leveraged effectively across various fields, ensuring both computational efficiency and accuracy.

As the development of Lunar Exploration Rover Systems (LERS) continues to advance, the need for precise and efficient real-time inverse kinematics (IK) solutions becomes increasingly critical, especially when multiple robotic arms must coordinate the handling of a shared payload. The ability of two or more arms to operate synchronously while managing complex tasks such as assembling infrastructure or manipulating lunar samples requires an IK solution that not only ensures accuracy but also operates with minimal computational overhead. The design of LERS equipped with dual arms, each capable of executing real-time trajectories while jointly manipulating a common load, is still in its early stages. However, developing an efficient closed-form analytical solution for the IK of each arm is essential for achieving the precision and speed necessary in these high-stakes, time-sensitive environments.

This paper presents a **computationally efficient closed-form analytical solution** for the inverse kinematics of LERS-mounted arms, addressing these critical requirements. Current IK methods often fall short when tasked with handling the real-time coordination of multiple robotic arms, particularly in complex and constrained environments like the lunar surface. This underscores a significant gap in the current state of robotics, pointing to the complexities of designing a robust

LERS capable of executing synchronized operations with multiple arms. The need for further research in this area is evident, as future lunar missions will increasingly rely on such systems to carry out coordinated, precision-based tasks with minimal delay and maximum reliability.

The structure of this paper is as follows: Section 2 covers the functional capabilities of the Lunar Exploration Rover System (LERS) and details its mission goals. In Section 3, we explore the system design of LERS, emphasizing the critical components that facilitate its autonomous operations. Section 4 is dedicated to discussing the closed-form analytical approaches for solving the inverse kinematics of the LERS arm, incorporating a geometric analytical technique. Finally, Section 5 provides a summary of the key findings and suggests directions for future research.

## II. Operational Capabilities and Mission Overview of LERS

The Lunar Exploration Rover System (LERS) marks a significant leap forward in lunar exploration, offering a wide range of capabilities designed to support various mission types (Fig. 1). Its adaptable design enables LERS to perform tasks that bolster both robotic and crewed missions, expanding the possibilities for lunar surface operations.

A key mission for LERS is to conduct scientific exploration and map the lunar terrain, where the rover traverses diverse landscapes to create detailed topographical maps. These high-resolution maps are crucial for planning future missions, charting safe paths, and pinpointing sites of scientific interest. Through autonomous mapping, LERS contributes to a deeper understanding of the Moon's geological features and helps identify areas that may hold valuable resources.

LERS is also central to sample collection and analysis, utilizing its robotic arm to gather lunar rock and soil from a variety of locations. The analysis of these materials sheds light on the Moon's origins, its development over time, and its potential to support future human activities. By conducting subsurface drilling and regular regolith collection, LERS provides essential data on the composition and potential resources of the lunar environment. Beyond scientific objectives, LERS is instrumental in construction and infrastructure tasks. Its robotic arm is specially designed to aid in building structures, installing equipment, and maintaining lunar habitats. These capabilities are crucial for establishing a sustainable human presence on the Moon, supporting efforts such as setting up outposts and assembling critical infrastructure.

LERS also is intended to operate autonomously, carrying out geological assessments, searching for water ice, and locating mineral resources without human intervention. Its autonomous systems will allow for ongoing exploration, even during communication delays or operator unavailability, ensuring continuous data collection over long durations.

Additionally, LERS will support real-time remote control, allowing mission teams on Earth to direct its movements and actions with precision. This feature is particularly important for performing delicate operations in challenging environments, such as deploying instruments or handling fragile equipment. LERS is also designed to integrate with international missions, working collaboratively with other space agencies to achieve shared exploration goals. Its ability to transport tools, set up equipment, and carry out joint tasks enhances global lunar exploration efforts and aids in the development of sustainable operations on the Moon.

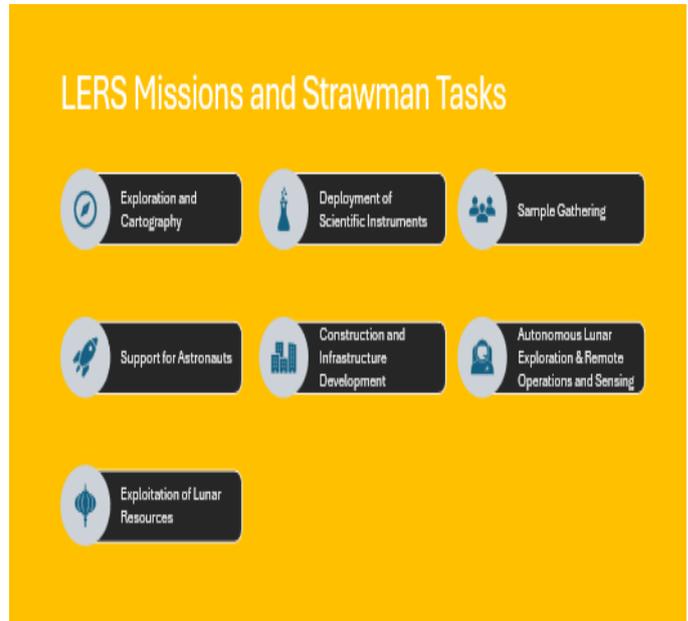

Fig. 1. LERS Mission Objectives and Operational Capabilities

## III. Overview of LERS Sytem Architecture

The Lunar Exploration Rover System (LERS) is engineered as a flexible platform capable of supporting a diverse array of lunar missions, with its architecture specifically designed to address the unique challenges of lunar exploration. This architecture is a well-coordinated framework consisting of multiple interconnected subsystems, each essential to the rover's efficiency and dependability. At the heart of LERS is an integrated system that combines both autonomous and remotely operated components, which interact through an advanced communication system to ensure seamless coordination between the rover's movement and the manipulation tasks performed by its robotic arm.

LERS's architecture is broadly organized into two primary subsystems: the remote subsystem and the local subsystem (Fig. 2). These components are linked through a central control unit, which facilitates decision-making and manages the flow of data. The remote subsystem is tasked with navigating the lunar surface, utilizing a suite of sensors, cameras, and real-time navigation algorithms to traverse difficult terrain. It also includes a force-sensitive gripper designed for precise environmental interaction, enabling the rover to carefully manipulate objects and carry out scientific experiments. The local subsystem incorporates simulation and control tools such as the Unity Simulator, Virtual Reality systems, and MATLAB/Simulink environments, along with hand-controllers for manual operation (Fig. 3).

A highly reliable communication network connects all subsystems, allowing continuous real-time data exchange between LERS and mission control. This architecture ensures that the rover can function autonomously while still enabling human oversight, even when subject to time delays. The modular design of the system makes it highly adaptable, allowing for future enhancements and the addition of new tools or sensors as mission objectives evolve.

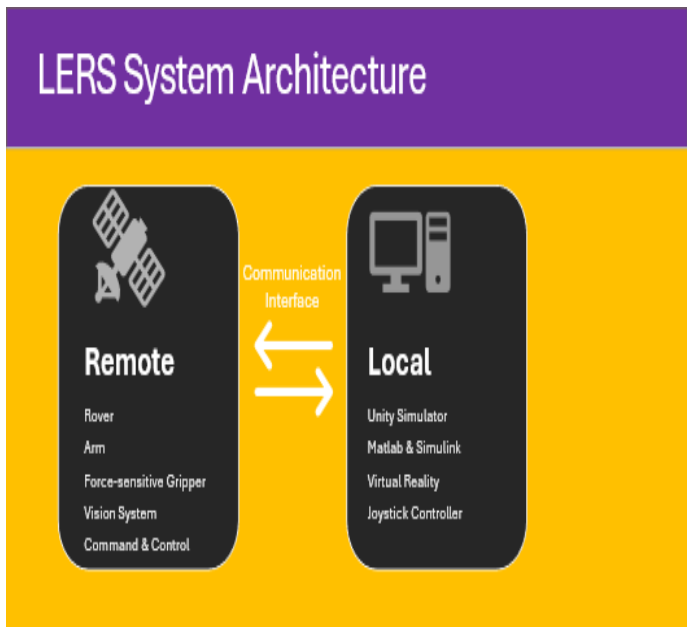

Fig. 2. LERS High-Level System Architecture

The creation of the Lunar Exploration Rover System (LERS) was guided by a systematic, iterative design approach, where each phase concentrated on enhancing both the mechanical aspects and software capabilities. Extensive simulations, combined with hands-on physical tests, were conducted to ensure that both the rover and its robotic arm could operate effectively under the extreme conditions of the lunar surface. From climbing steep inclines to handling delicate tasks in the Moon's low-gravity environment, the design was thoroughly evaluated. A testbed setup, replicating real lunar conditions, played a pivotal role in fine-tuning the system architecture. This setup was essential for validating the system's durability and confirming that all subsystems function cohesively to meet mission requirements.

Attached to the rover, the robotic arms are indispensable for carrying out precision operations, seamlessly integrating mechanical, electrical, and electronic systems, including a specialized Command, Control, and Communication ($C^3$) module. These arms are engineered to perform a range of tasks, from gathering lunar samples to deploying mission-critical equipment. Figs. 4a and 4b present the actual pictures, while Fig. 4c illustrates the CAD Model of the arm showcasing its design and functionality. The arm's structure features six Degrees of Freedom (DOF), allowing for flexible movement, and is powered by brushless DC gear motors. These motors are managed via a CAN Bus system, known for delivering accurate torque control and ensuring consistent, dependable performance, which is vital for scientific operations (Fig. 4d).

The J60 joint actuator offers exceptional torque performance while maintaining a lightweight design, making it an ideal component for robotic systems. With a peak torque-to-weight ratio of up to 56.48 Nm/kg (J60-10), it delivers robust power support in applications requiring efficient and powerful actuation. The actuator's compact dimensions are 76.5mm in length and 63mm in diameter, with a total weight of 480g.

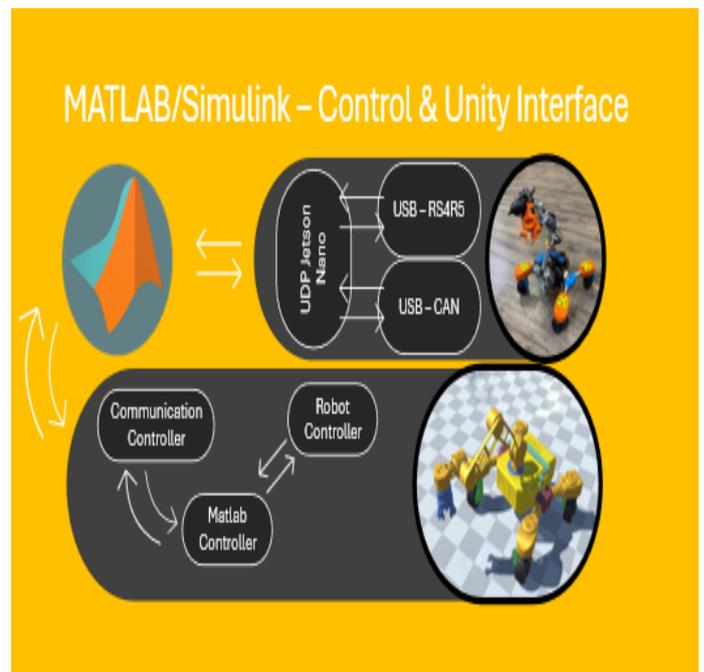

Fig. 3. MATLAB/Simulink™ – Control and Unity Interfaces

Electrically, the J60 operates within a voltage range of DC 18V to 36V, with a standard operating voltage of DC 24V. It is capable of handling a maximum motor phase current of 30A. In terms of torque, the J60 offers a peak torque of 19.94 Nm and a peak speed of 24.18 rad/s, with a joint torque constant of 0.8982 Nm/A. The maximum torque-to-weight ratio is 41.54 Nm/kg, making the J60 one of the most efficient actuators in its category.

For feedback and control, the J60 is equipped with an absolute value encoder with a resolution of 14 bits, ensuring precise position feedback. Communication is achieved through the CAN protocol, with a baud rate of 1 Mbps and a control frequency of 1 kHz, allowing for real-time control and synchronization in robotic systems.

The **CyberGear electric motor** is a high-performance actuator designed to deliver both exceptional torque and power density, making it well-suited for advanced robotic systems. The motor's torque density reaches up to **37.85 N·m/kg**, while the power density can achieve **511.04 W/kg**, ensuring efficient and powerful actuation in compact spaces. Weighing approximately **317g ± 3g**, this actuator provides an excellent balance of power and weight for robotic applications. The **CyberGear** motor operates at a rated voltage of **24V DC**, delivering a continuous torque of **4 N·m** and a peak torque of **12 N·m**. Its continuous torque maximum speed reaches **240 rpm ± 10%**, with a corresponding continuous phase current of **6.5A ± 10%**. The motor can handle a peak phase current of **23A ± 10%**, and its maximum speed is **296 rpm ± 10%**. The motor's torque constant is **0.87 N·m/A**, and it features a back electromotive force constant of **0.054 - 0.057 V/rpm.**

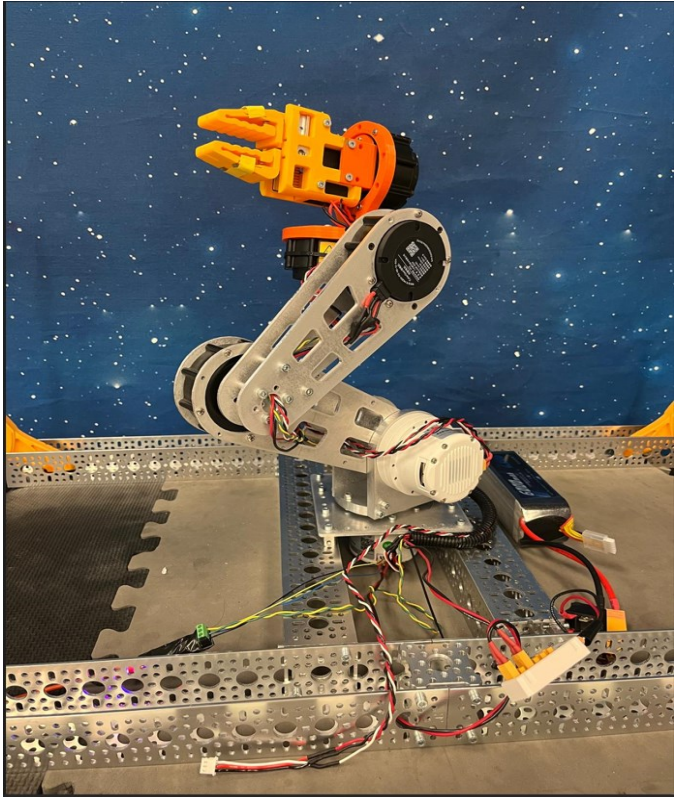

Fig. 4a. LERS Arm

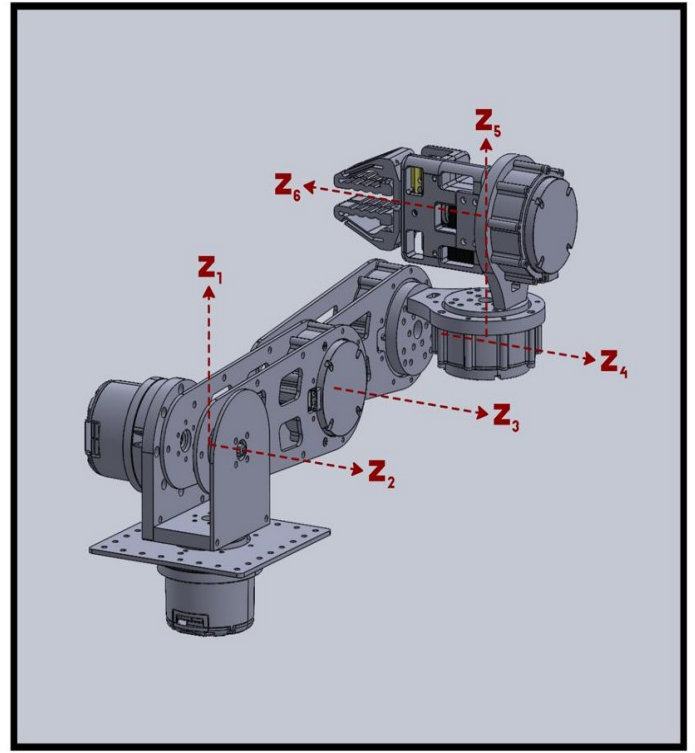

Fig. 4c. LERS Arm CAD Model

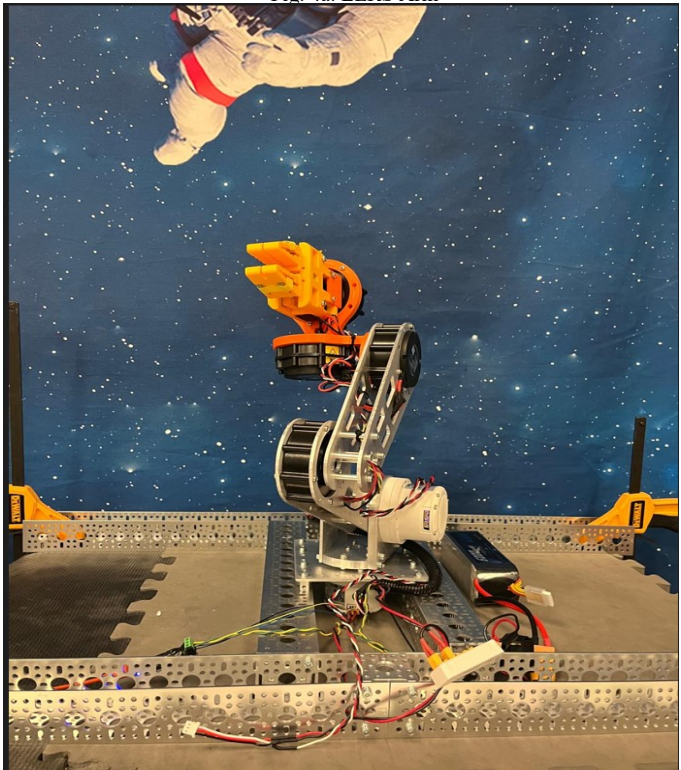

Fig. 4b. LERS Arm - A Different Perspective

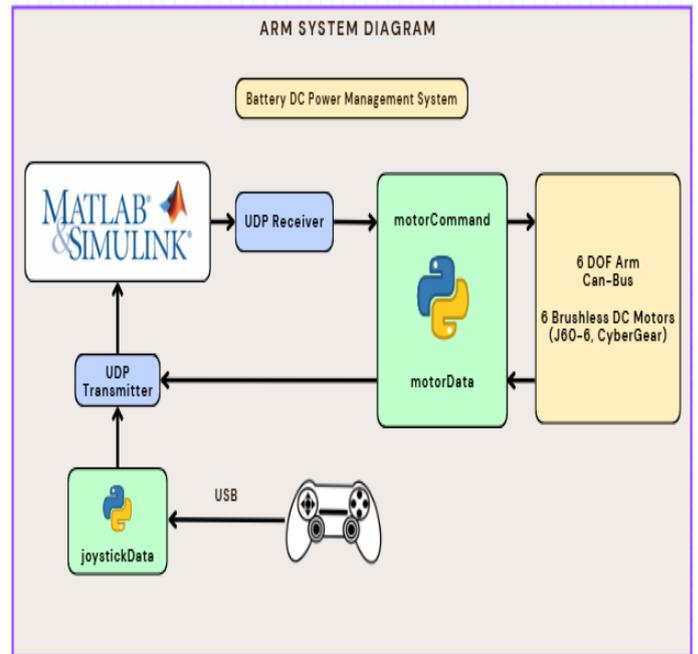

Fig. 4d. Can Bus Control Interface

Communication is facilitated through **CAN protocol** with a baud rate of **1 Mbps**, ensuring fast and reliable data transmission. The motor is equipped with a **14-bit absolute encoder**, allowing precise positional feedback and control.

A key element of the arm's design is the inclusion of 360-degree magnetic encoders, which provide real-time feedback on the servo's rotational position. This feedback loop is essential for maintaining precise control over the arm's movements, ensuring accuracy in delicate operations, especially during scientific data collection and experimentation. Additionally, the brushless motors offer programmable torque settings, enabling the arm to finely adjust its applied force, making it highly adaptable to the specific requirements of various tasks in a lunar environment.

## IV. INVERSE KINEMATICS MATHEMATICAL MODEL

This section introduces an approach for solving the inverse kinematics of the LERS arm, relying on geometric principles. This method is critical for determining the joint angles necessary to achieve a specific position and orientation of the robotic arm's end-effector, based on geometric principles. The **geometric approach** leverages the spatial relationships between different segments of the robotic arm.

By analyzing the geometric configuration of the arm's links and joints, this method provides a more intuitive, visual understanding of how the arm moves through space. This technique is particularly effective for systems with simpler kinematic structures, where the relationships between joint angles and end-effector position can be derived through straightforward trigonometric calculations. Geometric methods are often faster in certain contexts because they bypass the need for more complex matrix operations.

To illustrate this approach, Figure 5 presents the geometric configuration of the LERS arm, along with the assigned coordinate frames for each joint. This setup serves as the foundation for the geometric method, ensuring that all relevant spatial relationships and orientations are properly defined for the subsequent kinematic analysis.

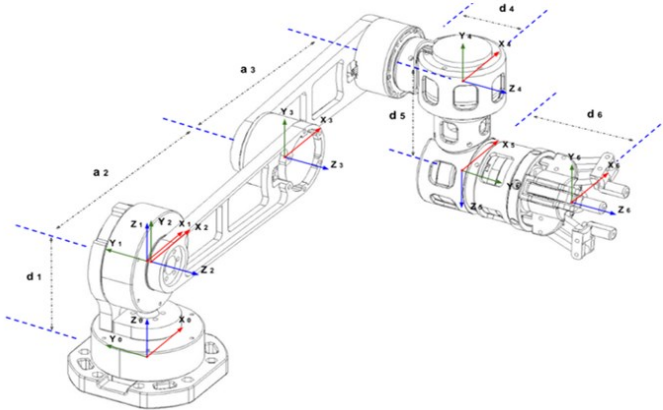

Fig. 5. 6 DOF LERS Robotics Arm l

We begin by defining the position vectors of the arm's end-effector and subsequently deduce the joint angles $\theta_i$, where $i$ varies from 1 to 6. The joint angles are defined by measuring rotation around the $z_i$ axis for each joint, starting from the home configuration. These measurements adhere to the right-hand rule for positive rotation, as illustrated in Figure 5.

The end-effector's position vector measured from the base coordinate frame $\{x_o, y_o, z_o\}$ is denoted as $P_6$ and is expressed as:

$$P_6 = \{x_6, y_6, z_6\} \quad (1)$$

For the fifth joint, the position vector $P_5$ is related to $P_6$ by the following equation:

$$P_5 = P_6 - d_6 \mathbf{k}_6 \quad (2)$$

where $d_6$ is the known distance along the $z_6$ axis, and $\mathbf{k}_6$ is the unit vector along the $z_6$ axis.

To consider the absence of offset $d_4$, we project $P_5$ onto the plane which encompasses the link designated as $a_2$ in Fig. 5, resulting in the derivation of $P'_5$.

$$P'_5 = P_5 - d_4 \mathbf{r}_0 \quad (3)$$

where $\mathbf{r}_0$ is the unit vector perpendicular to the plane containing link $a_2$ defined as:

$$\mathbf{r}_0 = \{\sin(\theta_1), -\cos(\theta_1), 0\} \quad (4)$$

Thus, the projected vector $P'_5$ is obtained, and its components can be written as:

$$x_{5p} = x_5 - d_4 \sin(\theta_1) \quad (5)$$
$$y_{5p} = y_5 + d_4 \cos(\theta_1) \quad (6)$$
$$z_{5p} = z_5 \quad (7)$$

From these components, we derive the relationship involving $tan(\theta_1)$ as follows:

$$\frac{y_{5p}}{x_{5p}} = \frac{y_5 + d_4 \cos(\theta_1)}{x_5 - d_4 \sin(\theta_1)} = \tan(\theta_1) = \frac{\sin(\theta_1)}{\cos(\theta_1)} \quad (8)$$

By equating the expression for $tan(\theta_1)$, we deduce the following:

$$d_4 = x_5 \sin(\theta_1) - y_5 \cos(\theta_1) \quad (9)$$

Reformulating the above equation, we establish:

$$K_1 \cos(\theta_1) + K_2 \sin(\theta_1) = K_3 \quad (10)$$

where $K_1 = -y_5$, $K_2 = x_5$ and $K_3 = d_4$.

We apply the half-angle substitution by letting $t = tan(\theta_1/2)$, which leads to:

$$\cos(\theta_1) = \frac{1-t^2}{1+t^2}, \quad \sin(\theta_1) = \frac{2t}{1+t^2} \quad (11)$$

This gives us a quadratic in $t$:

$$(K_3 + K_1)t^2 - 2K_2 t + (K_3 - K_1) = 0 \quad (12)$$

Solving for $t$, we find:

$$t = \frac{K_2 \pm \sqrt{K_2^2 - (K_3^2 - K_1^2)}}{K_3 + K_1} = \tan\left(\frac{\theta_1}{2}\right) \quad (13)$$

The angle $\theta_1$ is then:

$$\theta_1 = 2 \arctan\left(\frac{K_2 \pm \sqrt{K_2^2 - (K_3^2 - K_1^2)}}{K_3 + K_1}\right) \quad (14)$$

Alternatively, we can express $\theta_1$ in terms of α, where α is the angle between the x5 and y5 components of the position vector.

$$\frac{d_4}{\sqrt{x_5^2 + y_5^2}} = \frac{x_5}{\sqrt{x_5^2 + y_5^2}} \sin(\theta_1) - \frac{y_5}{\sqrt{x_5^2 + y_5^2}} \cos(\theta_1) \quad (15)$$

$$\frac{d_4}{\sqrt{x_5^2 + y_5^2}} = \sin(\theta_1 - \alpha) \quad (16)$$

Then,

$$\theta_1 - \alpha = \arcsin\left(\frac{d_4}{\sqrt{x_5^2 + y_5^2}}\right) \quad (17)$$

And,

$$\tan(\alpha) = \frac{y_5}{x_5} \quad (18)$$

We obtain:

$$\theta_1 = \arcsin\left(\frac{d_4}{\sqrt{x_5^2 + y_5^2}}\right) + atan2(y_5, x_5) \quad (19)$$

The angle $\theta_5$ is determined by the orientation relationship between vectors $\mathbf{r}_0$ and $\mathbf{k}_6$. The cosine and sine of $\theta_5$ can be found as:

$$\cos(\theta_5) = \mathbf{r}_0 \cdot \mathbf{k}_6, \quad \sin(\theta_5) = \|\mathbf{r}_0 \times \mathbf{k}_6\| \quad (20)$$

Therefore, $\theta_5$ is computed as:

$$\theta_5 = atan2(\|\mathbf{r}_0 \times \mathbf{k}_6\|, \mathbf{r}_0 \cdot \mathbf{k}_6) \quad (21)$$

The joint angles $\theta_2$, $\theta_3$, $\theta_4$ lie in the plane defined by links $a_2$ and $a_3$, and the projection of link $d_3$. The cumulative angle $\phi$ is given by:

$$\phi = \theta_2 + \theta_3 + \theta_4 - \frac{\pi}{2} \quad (22)$$

We determine the vector $\mathbf{k}_5$ as the cross product of $\mathbf{r}_0$ and $\mathbf{k}_6$:

$$\mathbf{k}_5 = \mathbf{r}_0 \times \mathbf{k}_6 \quad (23)$$

The angle $\phi$ can then be calculated as the angle between $\mathbf{k}_5$ and the local horizontal unit vector $\mathbf{r}_1$:

$$\mathbf{r}_1 = \{\cos(\theta_1), \sin(\theta_1), 0\} \quad (24)$$

The angle $\phi$ is given by:

$$\phi = atan2(\|\mathbf{r}_1 \times \mathbf{k}_5\|, \mathbf{r}_1 \cdot \mathbf{k}_5) \quad (25)$$

The new variables $X$ and $Z$ are expressed in terms of $x_{5p}$, $y_{5p}$, and $z_5$. Specifically, $X$ is defined as the magnitude of the projection on the x-y-plane, which can be represented mathematically as:

$$X = \sqrt{x_{5p}^2 + y_{5p}^2}$$

On the other hand, $Z$ is equated to $z_{5p}$, which is identical to $z_5$, implying that $Z = z_{5p} = z_5$.

$X$ and $Z$ are demonstrated in Figure 6 below. The X-axis represents the local horizontal axis, which is attached to the rotating frame and subject to rotation by $\theta_1$.

In the local plane coordinates, the expressions for $X$ and $Z$ are:

$$X + Zj = a_2 e^{(j\theta_2)} + a_3 e^{(j(\theta_3+\theta_4))} + d_5 e^{(j\phi)} \quad (26)$$

$$X - Zj = a_2 e^{(-j\theta_2)} + a_3 e^{(-j(\theta_3+\theta_4))} + d_5 e^{(-j\phi)} \quad (27)$$

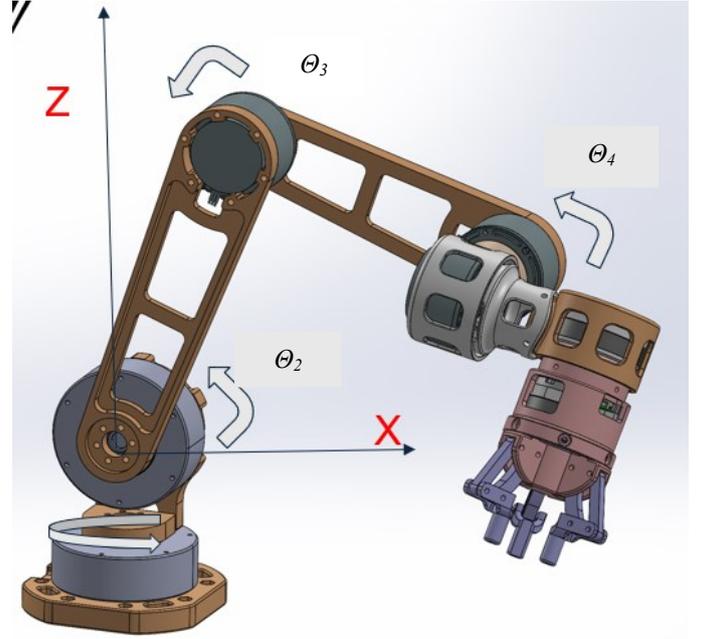

Fig. 6. Definitions of X and Z

Multiplying these equations yields a relation in the cosine and sine terms of $\theta_2$:

$$[2a_2 X - 2a_2 d_5 \cos(\phi)] \cos(\theta_2) + [2a_2 Z - 2a_2 d_5 \sin(\phi)] \sin(\theta_2) \quad (28)$$

$$= X^2 + Z^2 + a_2^2 + d_5^2 - a_3^2 - 2d_5 Z \sin(\phi) - 2d_5 X \cos(\phi). \quad (29)$$

This equation can further be written as:

$$L_1 \cos(\theta_2) + L_2 \sin(\theta_2) = L_3 \quad (30)$$

where $L_1$, $L_2$, and $L_3$ are defined as:

$$L_1 = 2a_2 X - 2a_2 d_5 \cos(\phi) \quad (31)$$
$$L_2 = 2a_2 Z - 2a_2 d_5 \sin(\phi) \quad (32)$$
$$L_3 = X^2 + Z^2 + a_2^2 + d_5^2 - a_3^2 - 2a_3 Z \sin(\phi) - 2d_5 \cos(\phi) \quad (33)$$

This can be simplified to a quadratic equation by expressing $\theta_2$ with half-angle substitutions in terms of $t = tan(\theta_2/2)$:

$$(L_3 + L_1)t^2 - 2L_2 t + (L_3 - L_1) = 0. \quad (34)$$

Solving for $t$ gives us:

$$t = \frac{L_2 \pm \sqrt{L_2^2 - (L_3^2 - L_1^2)}}{L_3 + L_1} = \tan(\frac{\theta_2}{2}), \quad (35)$$

and hence,

$$\theta_2 = 2\arctan\left(\frac{L_2 \pm \sqrt{L_2^2 - (L_3^2 - L_1^2)}}{L_3 + L_1}\right). \quad (36)$$

Given $\theta_2$ and the vector lengths, we define the following for convenience in the derivation of $\theta_3$:

$$L_4 = X - a_2 \cos(\theta_2) - d_5 \cos(\phi), \quad (37)$$
$$L_5 = a_3 \cos(\theta_2), \quad (38)$$
$$L_6 = a_3 \sin(\theta_2). \quad (39)$$

The trigonometric identities for $cos(\theta_3)$ and $sin(\theta_3)$ in terms of the half-angle tangent $t = tan(\theta_3/2)$ are:

$$\cos(\theta_3) = \frac{1-t^2}{1+t^2}, \quad (40)$$

$$\sin(\theta_3) = \frac{2t}{1+t^2}. \quad (41)$$

Solving for t using the quadratic formula gives:

$$t = \frac{-L_6 \pm \sqrt{L_6^2 - (L_4^2 - L_5^2)}}{L_4 + L_5} = \tan\left(\frac{\theta_3}{2}\right). \quad (42)$$

Therefore, the angle $\theta_3$ is computed as:

$$\theta_3 = 2\arctan\left(\frac{-L_6 \pm \sqrt{L_6^2 - (L_4^2 - L_5^2)}}{L_4 + L_5}\right). \quad (43)$$

The angle $\theta_4$ is derived from the cumulative angle $\phi$, which accounts for the sum of $\theta_2$, $\theta_3$, and $\theta_4$, adjusted by $-\pi/2$. We find $\theta_4$ by subtracting the known values of $\theta_2$ and $\theta_3$ from $\phi$:

$$\theta_4 = \phi - (\theta_2 + \theta_3) + \frac{\pi}{2}. \quad (44)$$

To compute $\theta_6$, the angle between the unit vectors $\mathbf{m}_5$ and $\mathbf{m}_6$, we use the dot product for $cos(\theta_6)$ and the magnitude of the cross product for $sin(\theta_6)$:

$$\cos(\theta_6) = \mathbf{m}_5 \cdot \mathbf{m}_6, \quad (45)$$

$$\sin(\theta_6) = \|\mathbf{m}_5 \times \mathbf{m}_6\|. \quad (46)$$

where $\mathbf{m}_5$ and $\mathbf{m}_6$ are the unit vectors along the $\mathbf{x}_5$ and $\mathbf{x}_6$, respectively. The angle $\theta_6$ is then given by:

$$\theta_6 = \arctan 2(\|\mathbf{m}_5 \times \mathbf{m}_6\|, \mathbf{m}_5 \cdot \mathbf{m}_6). \quad (47)$$

where $\mathbf{k}_5 = \mathbf{r}_0 \times \mathbf{k}_6$ and $\mathbf{m}_5 = \mathbf{k}_6 \times \mathbf{k}_5$. The vectors $\mathbf{m}_5$ and $\mathbf{k}_5$ are the unit vectors along the axes $\mathbf{x}_5$ and $\mathbf{z}_5$, respectively.

Figure 7 demonstrates the high accuracy of the analytical solution. The circles represent the exact positions of the end-effector, while the crosses (x) indicate the positions calculated using the joint angles obtained from the analytical inverse kinematics. The close alignment between the two sets of points highlights the excellent agreement. The exact locations are randomly selected within the robotic arm's reach.

I. CONCLUSION

The development of a closed-form analytical solution for the inverse kinematics of the 6 DOF serial robotic manipulator used in the Lunar Exploration Rover System (LERS) represents a significant leap forward in the field of space robotics. This solution enables the precise control of the arm's 6 revolute joints, allowing it to perform a wide range of critical tasks on the lunar surface, from delicate sample collection to the assembly of infrastructure needed for sustained human presence. The geometric approach used in this study offers a streamlined and computationally efficient method for solving the inverse kinematics problem, providing exact joint configurations without the iterative processes and high computational demands typical of numerical methods.

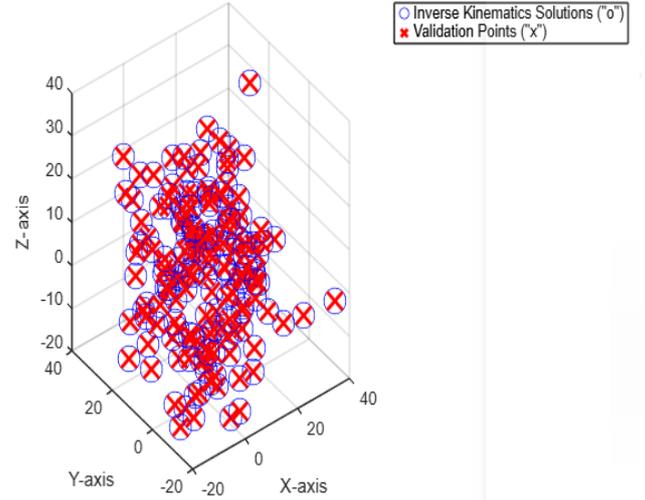

Fig. 7. Validation of Inverse Kinematics

The solution's effectiveness in handling real-time operations makes it particularly valuable for space missions, where time delays and limited computational resources pose significant challenges. The ability to operate autonomously in complex, unpredictable environments like the Moon is crucial for advancing space exploration, especially in the context of the ARTEMIS program, which aims to establish a sustainable human presence on the lunar surface. This IK solution also opens up new possibilities for multi-robot systems, where multiple robotic arms could work in coordination to manage shared payloads or execute highly synchronized tasks.

The findings presented in this paper have broader applications beyond lunar exploration. They are relevant to future space missions that require reliable and precise robotic systems for satellite servicing, space debris removal, and in-orbit assembly. As space exploration evolves, the integration of advanced robotic systems, such as the LERS arm with its 6 DOF capabilities, will become increasingly important for the success of both crewed and robotic missions. This research highlights the need for continued innovation in robotic control systems to ensure that space missions can be conducted safely, efficiently, and with minimal human intervention.